\documentclass[sigconf]{acmart}
\AtBeginDocument{%
  \providecommand\BibTeX{{%
    \normalfont B\kern-0.5em{\scshape i\kern-0.25em b}\kern-0.8em\TeX}}}

\copyrightyear{2022}
\acmYear{2022}
\setcopyright{acmlicensed}\acmConference[GECCO '22 Companion]{Genetic
and Evolutionary Computation Conference Companion}{July 9--13,
2022}{Boston, MA, USA}
\acmBooktitle{Genetic and Evolutionary Computation Conference Companion
(GECCO '22 Companion), July 9--13, 2022, Boston, MA, USA}
\acmPrice{15.00}
\acmDOI{10.1145/3520304.3533949}
\acmISBN{978-1-4503-9268-6/22/07}

\usepackage{mathtools}
\usepackage[utf8]{inputenc}
\usepackage{multirow}
\usepackage{url}

\usepackage{subcaption}
\usepackage{framed,multirow}
\usepackage{latexsym}
\usepackage[utf8]{inputenc}

\usepackage{algorithm}
\usepackage[noend]{algpseudocode}
\usepackage{setspace}
\let\Algorithm\algorithm
\renewcommand\algorithm[1][]{\Algorithm[#1]\setstretch{0.8}}

\algrenewcommand{\algorithmiccomment}[1]{\hskip3px$\#$ #1}
\algdef{SE}[SUBALG]{Indent}{EndIndent}{}{\algorithmicend\ }%
\algtext*{Indent}
\algtext*{EndIndent}
\usepackage{url}
\usepackage{xcolor}
\definecolor{newcolor}{rgb}{.8,.349,.1}
\usepackage{graphicx}

\usepackage{graphicx}

\newcommand{\bb}{\bf\large}

\begin{document}

\title{Evolution of Activation Functions for Deep Learning-Based Image Classification}

\author{Raz Lapid and Moshe Sipper}

\email{sipper@bgu.ac.il} 
\authornotemark[1]
\orcid{0000-0003-1811-472X}
\affiliation{%
  \institution{Department of Computer Science, Ben-Gurion University}
  \streetaddress{}
  \city{Beer Sheva 84105}
  \state{}
  \country{Israel}
  \postcode{84105}
}

\renewcommand{\shortauthors}{Lapid and Sipper}

\begin{abstract}
Activation functions (AFs) play a pivotal role in the performance of  neural networks.
The Rectified Linear Unit (ReLU) is currently the most commonly used AF. Several replacements to ReLU have been suggested but improvements have proven inconsistent. Some AFs exhibit better performance for specific tasks, but it is hard to know a priori how to select the appropriate one(s). Studying both standard fully connected neural networks (FCNs) and convolutional neural networks (CNNs), we propose a novel, three-population, coevolutionary algorithm to evolve AFs, and compare it to four other methods, both evolutionary and non-evolutionary. 
Tested on four datasets---MNIST, FashionMNIST, KMNIST, and USPS---coevolution proves to be a performant algorithm for finding good AFs and AF architectures.
\end{abstract}

\ccsdesc[500]{Computing methodologies~Neural networks}
\ccsdesc[500]{Computing methodologies~Image processing}
\ccsdesc[500]{Theory of computation~Evolutionary algorithms}

\keywords{deep learning, activation functions, coevolution}

\maketitle

\section{Introduction} 
\label{sec:intro}
Artifical Neural Networks (ANNs), and, specifically, Deep Neural Networks (DNNs), have gained much traction in recent years and are now being effectively put to use in a variety of applications. Considerable work has been done to improve training and testing performance, including various initialization techniques, weight-tuning algorithms, different architectures, and more. However, one hyperparameter is usually left untouched: the activation function (AF). While recent work has seen the design of novel AFs \cite{agostinelli2014learning,saha2019evolution,sharma2017activation,sipper2021neural}, the Rectified Linear Unit (ReLU) remains by far the most commonly used one, mainly due to its overcoming the vanishing-gradient problem, thus affording faster learning and better performance.

AFs are crucial first and foremost because they transform a neural network from a simple linear algorithm into a non-linear one, thus allowing the network to learn complex mappings and functions. If the AFs of a neural network were removed, the whole architecture could be reduced to a linear operation on its input, with very limited use. Traditionally, the most commonly used AFs were sigmoid functions---such as logistic and hyperbolic tangent---which are bounded, differentiable, and monotonic. Such functions are prone to the vanishing-gradient problem \cite{hochreiter1998vanishing}: when the function value is either too high or too low, the derivative becomes very small, and learning becomes very poor. To address this issue, different functions have been developed, the most prominent of which being the ReLU. 
Though it has proven highly performant, the ReLU is susceptible to a problem known as ``dying'' \cite{lu2019dying}: because all negative values are mapped to zero, there might be a scenario where a large number of ReLU neurons only output a 0 value. It might cause the entire network to ``die'', resulting in a constant function. 

Different variants of ReLU, such as PReLU \cite{zhang2018multiple} and Leaky ReLU \cite{zhang2017dilated}, have been proposed to address this issue. Although there has been much research towards creating different variants, ReLU remains the most commonly used AF. 
Highly popular architectures---including AlexNet, ZFNet, VGGNet, SegNet, GoogLeNet, SqueezeNet, ResNet, ResNeXt, MobileNet, and SeNet---have a common AF setup, with ReLUs in the hidden layers and a softmax function in the output layer \cite{nwankpa2018activation}. 

This paper introduces a novel coevolutionary algorithm to evolve AFs for image-classification tasks. Our method is able to handle the simultaneous coevolution of three types of AFs: input-layer AFs, hidden-layer AFs, and output-layer AFs. We surmise that combining different AFs throughout the architecture may improve the network's performance. 
We compare our novel algorithm to four different methods: standard ReLU- or LeakyReLU-based networks, networks whose AFs are produced randomly, and two forms of single-population evolution, differing in whether an individual represents a single AF or three AFs. We chose ReLU and LeakyReLU as baseline AFs since we noticed that they are the most-used functions in the deep-learning domain.

In the next section we review the literature on evolving new AFs through evolutionary computation. Section~\ref{sec:methods} delineates the different methods used in this paper and our proposed coevolutionary algorithm. The experimental setup used to test the methods is delineated in Section~\ref{sec:exp}.
We present results and analyze them in Section~\ref{sec:res}. 
Finally, we offer concluding remarks in Section~\ref{sec:remarks}.

\section{Previous Work} 
\label{sec:prevWork}
There has been extensive research into using evolutionary algorithms for optimizing and improving ANNs and DNNs, which can be divided into two main categories: 
1) topology optimization \cite{miikkulainen2019evolving}, and 
2) hyperparameter and weight optimization \cite{8489520}.  
Topology-optimization methods can be further divided into two groups: constructive and destructive. Constructive techniques start with a simple topology and gradually enhance its complexity until it meets an optimality requirement. 
Contrarily, destructive methods start with a complicated topology and gradually remove unneeded components. 

One of the most well-known methods is NeuroEvolution of Augmenting Topologies (NEAT) \cite{stanley2002evolving}. This is an evolutionary algorithm-based method for developing neural networks. NEAT follows the constructive approach, with the evolutionary algorithm beginning with small, basic networks and gradually increasing their complexity over the generations. Finding intricate and complex neural networks is possible through this iterative approach. NEAT tries to modify network topologies and weights in an attempt to strike a balance between the fitness of developed solutions and their diversity.

\cite{sun2019evolving} used a genetic algorithm to evolve neural-network architectures and weight initialization of deep CNNs for image-classification problems. Another work that used evolution to search for neural-network architectures is \cite{9672175}, in which they utilized a partial weight-sharing, one-shot neural architecture search framework that directly evolved complete neural-network architectures. \cite{10.1145/3491396.3506510} evolved neural-network architectures using a training-free genetic algorithm.

There are many more papers that aim to optimize a neural network's architecture without evolution, e.g., \cite{Franchini2023}, wherein they introduced an automatic machine learning technique to optimise a CNN's architecture by predicting the performance of the network. Another interesting work is \cite{DING2022}, in which they trained an over-parameterized network and then used a pruning criterion to prune the network. 

Hyperparameter optimization is also a thriving field of research. The impact of hyperparameters on different deep-learning architectures has revealed complicated connections, with hyperparameters that increase performance in basic networks not having the same effect with more sophisticated topologies \cite{breuel2015effects}. Selecting the right AF(s) for a specific task is crucial for the success of the neural network. 

There are four commonly used methods for hyperparameter selection (including AFs): 
1) manual search, 
2) grid search, 
3) random search, 
and
4) Bayesian optimization. 
Manual search simply refers to the user's manually selecting hyperparameters,
a method often used in practice because it is quick and can result in sufficing results. Nonetheless, this technique makes it difficult to replicate results on new data, especially for non-experts. 
Grid search exhaustively generates sets of hyperparameters from values supplied by the user.
This approach relies on the programmer's knowledge of the problem and produces repeatable results, but it is inefficient when exploring a large hyperparameter space. Grid search is commonly used because it is simple to set up, parallelize, and explore the whole search space (to some extent). Random search is similar to grid search except that hyperparameter values are chosen at random from ranges specified by the user; it often works better than other methods \cite{bergstra2012random}. Bayesian optimization is based upon using the information gained from a given trial experiment to decide how to adjust the hyperparameters for the next trial \cite{snoek2012practical}.

A recent work by \cite{Nader2021} focused on evolving AFs for neural networks. Their work differs in several aspects from our novel coevolutionary algorithm:
\begin{enumerate}
    \item They used standard evolution, whereas we use coevolution.
    \item We deploy Cartesian genetic programming (CGP)---with its immanent bloat-control---rather than (bloat-susceptible) tree-based GP.
    \item We use evolved AFs in all network layers, as opposed to their work, which only used evolved AFs in the hidden layers.
    \item We use four benchmark image datasets in full, whereas they used a small subset of image datasets, and tabular datasets.
    \item We investigate two different architectures (fully connected neural network and convolutional neural network), with two different baseline AFs (ReLU and Leaky ReLU), while they used a randomly chosen architecture for each of the tested datasets.
\end{enumerate}

A number of papers evolve AFs using genetic programming \cite{basirat2018quest,10.1145/3377930.3389841, bingham2022discovering}, but we have not found any using \textit{coevolution} for this purpose. We believe coevolution is well-adapted to the problem at hand, given the different natures of AFs depending on their placement---input, hidden, or output layer.

We are interested in coevolving AFs for fixed-topology neural networks. The coevolutionary algorithm iteratively evolves better AFs, a technique that has received less attention so far within the scope of AFs. 

\section{Methods} 
\label{sec:methods}
We compare five different methods for obtaining neural networks that perform a given task:
\begin{enumerate}
    \item Standard fully connected neural network (FCN) and standard convolutional neural network (CNN).
    \item Random FCN and random CNN.
    \item Single-population evolution, where an individual represents a single AF.
    \item Single-population evolution, where an individual represents three AFs: input-layer AF, hidden-layer AF, and output-layer AF.
    \item Coevolutionary algorithm, with three populations: a population of input-layer AFs, a population of hidden-layer AFs, and a population of output-layer AFs.
\end{enumerate}

We used the following hyperparmaters for the learning algorithm in each of the methods: learning rate -- $0.01$, optimizer -- Adam, batch size -- $\frac{1}{10}$ the size of the training set. PyTorch's default initialization was used both for weights and biases, meaning that values were initialized from $U(-\sqrt{k}, \sqrt{k})$, where $k=\frac{1}{number\_of\_input\_features}$.

Below we delineate each of the above methods in more detail. We used the same number of candidate solutions in each of the different methods to conduct a fair comparison. Also, before moving on to describe the evolutionary-based algorithms, we provide a brief description of Cartesian genetic programming, the flavor of evolutionary algorithm we used herein. 

\subsection{Standard neural networks}
\label{subsec:standard}
A standard FCN is typically composed of several building blocks that are chained sequentially, starting with data input and ending with an output (upon which a loss function is computed). In the \textit{Standard-FCN} architecture we used, the non-linear functions \texttt{torch.nn.ReLU} or \texttt{torch.nn.LeakyReLU} follow each fully connected layer (and before the final softmax layer).
In the \textit{Standard-CNN} architecture we used, \texttt{torch.nn.ReLU} or \texttt{torch.nn.LeakyReLU} follow each convolution layer and the last fully connected layer (before the final softmax layer).
ReLU and Leaky ReLU thus form our baseline AFs for comparison.

In our setting (for all methods), a network requires three AFs (which can be identical---or not):
an input-layer AF, a hidden-layer AF, and an output-layer AF.
For FCNs, the first AF is applied after the first \texttt{torch.nn.Linear} layer, the second AF is applied after each of the hidden \texttt{torch.nn.Linear} layers, and the third AF is applied  after the last \texttt{torch.nn.Linear} layer (before the softmax layer; see Table~\ref{tab:ann_params}). 
For CNNs, the first AF is applied after the first \texttt{torch.nn.Conv2D} layer, the second AF is applied after the second \texttt{torch.nn.Conv2D} layer, and the third AF is applied  after the first \texttt{torch.nn.Linear} layer (see Table~\ref{tab:cnn_params}).

\begin{table}
    \centering
    \caption{FCN architecture.}
    \label{tab:ann_params}
    \resizebox{0.45\textwidth}{!}{
    \begin{tabular}{|c|c|c|}
        \hline
         \textbf{Layer} & \textbf{Layer type} & \textbf{Number of neurons} \\ 
          \hline
          1 (input) & Fully Connected & 784 \\
          2-6 (hidden) & Fully Connected ($\times$ 5) & 32 \\
          7 (output) & Fully Connected & 10 \\
          \hline
    
    \end{tabular}
    }
    \vspace*{0.5 cm}
    \centering
    \caption{CNN architecture.}
    \label{tab:cnn_params}
    \resizebox{0.45\textwidth}{!}{
    \begin{tabular}{|c|c|c|c|c|}
        \hline
         \textbf{Layer} & \textbf{Layer type} & \textbf{No. channels} & \textbf{Filter size} & \textbf{Stride}\\ 
          \hline
          1 & Convolution & 32 & $3 \times 3$ & 1 \\
          2 & Max Pooling & N/A & $3 \times 3$ & 2 \\
          3 & Convolution & 64 & $3 \times 3$ & 1 \\
          4 & Max Pooling & N/A & $3 \times 3$ & 2 \\
          5 & Dropout $(p=0.5)$ & N/A & N/A & N/A \\
          6 & Fully Connected & 1600 & N/A & N/A \\
          7 & Fully Connected & 1000 & N/A & N/A \\
          8 & Fully Connected & 10 & N/A & N/A \\
          \hline

    \end{tabular}
    }
\end{table}

\subsection{Random neural networks}
Random search is a class of numerical optimization methods that do not consider the problem's gradient, allowing them to be utilized on functions that are neither continuous nor differentiable.
Herein, we randomly created a set of candidate (AF) solutions, with each one comprising three ordered AFs. We generated \textit{max\_iter}*3 candidate solutions for fair comparison with CGP (\textit{max\_iter}---see Table~\ref{tab:cgp_params}), by repeatedly creating a random initial generation through CGP. Each candidate solution was then trained over the train$_1$ set for 3 epochs\footnote{3 epochs proved a good tradeoff between evolution-driving fitness and runtime.} and its score was computed as the accuracy over the train$_2$ set. Then, we pick the best (top-accuracy) candidate solution for comparison.

\subsection{Cartesian genetic programming}
\label{subsec:cgp}
Cartesian genetic programming (CGP) is an evolutionary algorithm wherein an evolving individual is represented as a two-dimensional grid of computational nodes---often an a-cyclic graph---which together express a program \cite{10.1145/1388969.1389075}. It originally grew from a mechanism for developing digital circuits \cite{miller1997designing}. An individual is represented by a linear genome, composed of integer genes, each encoding a single node in the graph, which represents a specific function. 
A node consists of a function, from a given table of functions, and connections, specifying  where the data for the node comes from.

\begin{figure*}
    \centering
    \includegraphics[scale=0.55]{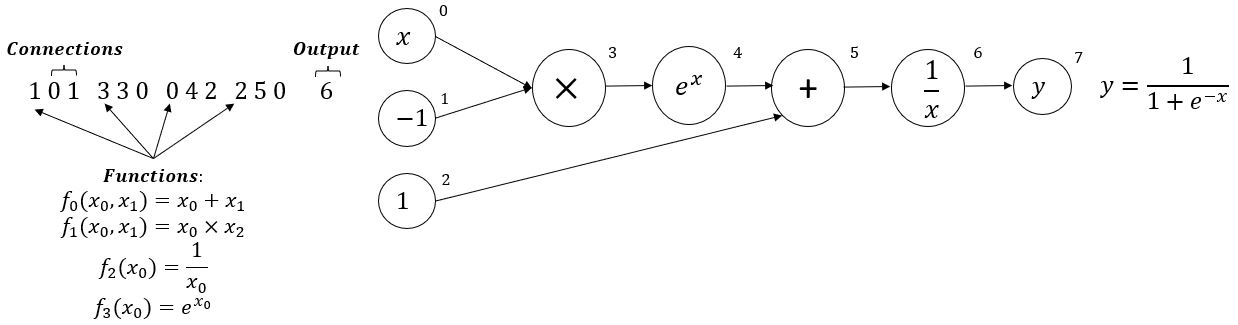}
    \caption{A sample CGP individual, with 3 inputs---$x, -1, 1$---and 1 output---$y$. 
    The genome consists of 5 3-valued genes, per 4 functional units, plus the output specification (no genes for the inputs). 
    The first value of each 3-valued gene is the function's index in the lookup table of functions (bottom-left), and the remaining two values are parameter nodes. The last gene determines the outputs to return. 
    In the example above, with $n_i$ representing node $i$: 
    node 3, gene 101, $f_1(n_0, n_1)=n_0 \times n_1$;
    node 4, gene 330, $f_3(n_3)=e^{n_3}$ (unary function, 3rd gene value ignored);
    node 5, gene 042, $f_0(n_4,n_2)=n_4 + n_2$; 
    node 6, gene 250, $f_2(n_5)=\frac{1}{n_5}$;
    node 7, output node, $n_6$ is the designated output value.
    The topology is fixed throughout evolution, while the genome evolves.}
    \label{fig:ind_example}
\end{figure*}

A sample individual is shown in Figure~\ref{fig:ind_example}. 
There are three hyperparameters that define the connectivity and dimensionality of the encoded architecture: 1) number of rows, 2) number of columns, and 3) number of rows to look back for connections---this limits the columns from which a node can acquire its inputs. This representation is simple, flexible, and convenient for many problems. Typically, evolution begins with a population of randomly selected candidate solutions. In our case, because we are evolving AFs, an individual in a population represents an AF. Driven by a fitness function and using stochastic modification operators, CGP is able to produce successively better models in an iterative manner. 

The CGP library used in this paper \cite{Ohjeah} evolves the population in a $(1 + \lambda)$-manner, i.e., in each generation it creates $\lambda$ offspring (we used the default $\lambda =4$) and compares their fitness to the parent individual. The fittest individual carries over to the next generation; in case of a draw, the offspring is preferred over the parent. Tournament selection is used, with tournament size $k=|population|$, single-point mutation, and no crossover. 
For more details, see \cite{miller2006redundancy,Ohjeah}.
Table~\ref{tab:cgp_params} delineates the hyperparameters used by the CGP algorithm, and Table~\ref{tab:primitives_list} shows the primitive set.

\begin{table}
    \centering
    \caption{Hyperparameters used by CGP.}
    \label{tab:cgp_params}
    \resizebox{0.45\textwidth}{!}{
    \begin{tabular}{r|c|l}
         \textbf{Parameter} & \textbf{Description} & \textbf{Value}\\ 
          \hline
          \textit{operators} & list of primitives & see Table~\ref{tab:primitives_list} \\
          \textit{n\_const} & number of symbolic constants & 0 \\
          \textit{n\_rows} & number of rows in the code block & 5 \\
          \textit{n\_columns} & number of columns in the code block & 5 \\
          \textit{n\_back} & number of rows to look back for connections & 5 \\
          \textit{n\_mutations} & number of mutations per offspring & 3 \\
          \textit{mutation\_method} & specific mutation method & point \\
          \textit{max\_iter} & maximum number of generations & 50\\
          \textit{lambda} & number of offspring per generation & 4 \\
          \textit{f\_tol} & absolute error acceptable for convergence & 0.01 \\
          \textit{n\_jobs} & number of jobs & 1 \\
    \end{tabular} 
    }
\end{table}

\begin{table}
    \centering
    \caption{Set of primitives used by CGP. Left: Standard AFs. Right: Mathematical functions.}
    \label{tab:primitives_list}
    \vspace{-8pt}
    \resizebox{0.45\textwidth}{!}{
    \begin{tabular}{r|l|r|l}
     \textbf{AF} & \textbf{Expression} & \textbf{AF} & \textbf{Expression} \\ 
      \hline
      ReLU & $f(x)=max(x,0)$ & Max & $f(x, y) = max(x, y)$ \\
      \hline
      Tanh & $f(x)=\frac{e^x - e^{-x}}{e^x + e^{-x}}$ & Min & $f(x, y) = min(x, y)$ \\
      \hline
      Leaky ReLU & $f(x<0) = negative\_slope \cdot x$ & Add & $f(x, y) = x + y$ \\ 
                          & $f(x\geq 0)=x$ & & \\
      \hline
      ELU & $f(x\leq 0) = (exp(x)-1)$ & Sub & $f(x, y) = x - y$  \\ 
                   & $f(x>0)=x$  & & \\
      \hline
      HardShrink & $f(x < -\lambda) = x$ & Mul & $f(x, y) = x \cdot y$ \\
                          & $f(x>\lambda)=x$ & & \\ 
                          & $f(-\lambda \leq x \leq \lambda)=0$ & & \\
      \hline
      CELU & $f(x) = max(0, x) + min(0, \alpha (e^{\frac{x}{\alpha}} - 1)$ & & \\ 
      \hline
      Hardtanh & $f(x > 1) = 1$ & & \\ 
                        & $f(x < -1)= -1$ & &  \\
                        & $f(-1 \leq x \leq  1)=x$ & & \\ 
      \hline
      Hardswish & $f(x \leq -3) = 0$ & & \\ 
                         & $f(x \geq 3)= -1$ & & \\ 
                         & $f(-3 < x \leq  3)=x\cdot \frac{(x+3)}{6}$ & & \\
      \hline
      Softshrink & $f(x > \lambda) = x - \lambda$ & & \\ 
                         & $f(x < - \lambda)= x + \lambda$ & & \\ 
                         & $f(- \lambda \leq x \leq  \lambda )=0$ & & \\
      \hline
      RReLU & $f(x \geq 0) = x$ & & \\ 
                     & $f(x < 0)= \alpha \cdot x$ & & \\
                     & ($\alpha$ is uniformly, randomly sampled) & & \\
      \hline
    \end{tabular} 
    }
\end{table}

\subsection{Evo-Single: Single-population evolution}
\begin{figure}[htp]
\centering
    \includegraphics[width=0.4\textwidth]{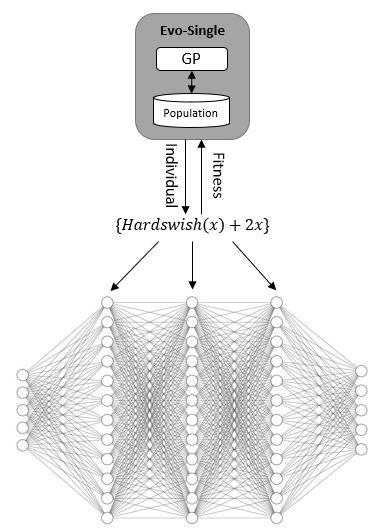}\hfill
    \caption{Evo-Single: Fitness computation of an individual in the population. The same AF is used throughout the network, in this example $Hardswish(x)+2x$. }
    \label{fig:evo-single}
\end{figure}

Evo-Single is a single-population evolutionary algorithm, wherein an individual represents a single AF used throughout the whole network (as explained in Section~\ref{subsec:standard}). 
We start with the ReLU AF or the Leaky ReLU AF. Then we iterate, mutating the AF at every iteration of the CGP algorithm. Each individual is then trained over the train$_1$ set for 3 epochs and its fitness is computed as the accuracy over the train$_2$ set. After $\textit{max\_iter}*3=150$ iterations, we select the individual whose performance on the train$_2$ set was best. This individual is then compared against the other methods. 
The fitness-computation scheme is shown in Figure~\ref{fig:evo-single}.

\subsection{Evo-Triple: Single-population evolution}
\begin{figure}[htp]
\centering
    \includegraphics[width=0.4\textwidth]{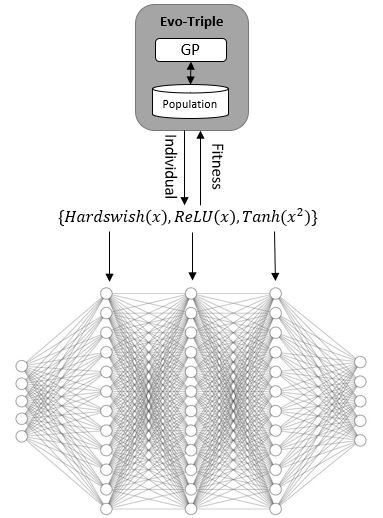}\hfill
    \caption{Evo-Triple: Fitness computation of an individual in the population, which comprises 3 AFs, in this example: $\{Hardswish(x), ReLU(x), Tanh(x^2)\}$.}
     \label{fig:evo-triple}
\end{figure}

Evo-Triple is a single-population evolutionary algorithm, wherein an individual comprises an array of 3 AFs, the first being the input-layer AF, the second being the hidden-layer AF, and the third being the output-layer AF. The AFs are used throughout the network as explained in Section~\ref{subsec:standard}. We start with 3 ReLU AFs or 3 Leaky ReLU AFs. Then we iterate, mutating each of the AFs at every iteration of the CGP algorithm. Each individual is then trained over the train$_1$ set for 3 epochs and its fitness is computed as the accuracy over the train$_2$ set. After $\textit{max\_iter}*3=150$ iterations, we select the individual whose performance on the train$_2$ set was best. This individual is than compared against the other methods. 
As with Evo-Single, training uses the train$_1$ set, and fitness is computed over the train$_2$ set. The fitness-computation scheme is shown in Figure~\ref{fig:evo-triple}.

\subsection{Coevo: Three-population coevolution}
\label{subsec:co-evo}

Coevolution refers to the simultaneous evolution of two or more species with coupled fitness \cite{Pena:2001}.
Coevolution can be mutualistic, parasitic, or commensalistic \cite{sipper2019solution}. In mutualistic coevolution, two or more species reciprocally affect each other beneficially. Parasitic coevolution is a competitive relationship between species. Commensalistic coevolution is a relationship between species where one of the species gains benefits while the other neither benefits nor is harmed. In this paper we focus on mutualistic, or cooperative coevolution, which is based on the collaboration of several independently evolving species to solve a problem. An individual's fitness is determined by its ability to cooperate with members of the other species. 

\begin{figure*}
\centering
    \includegraphics[scale=0.55]{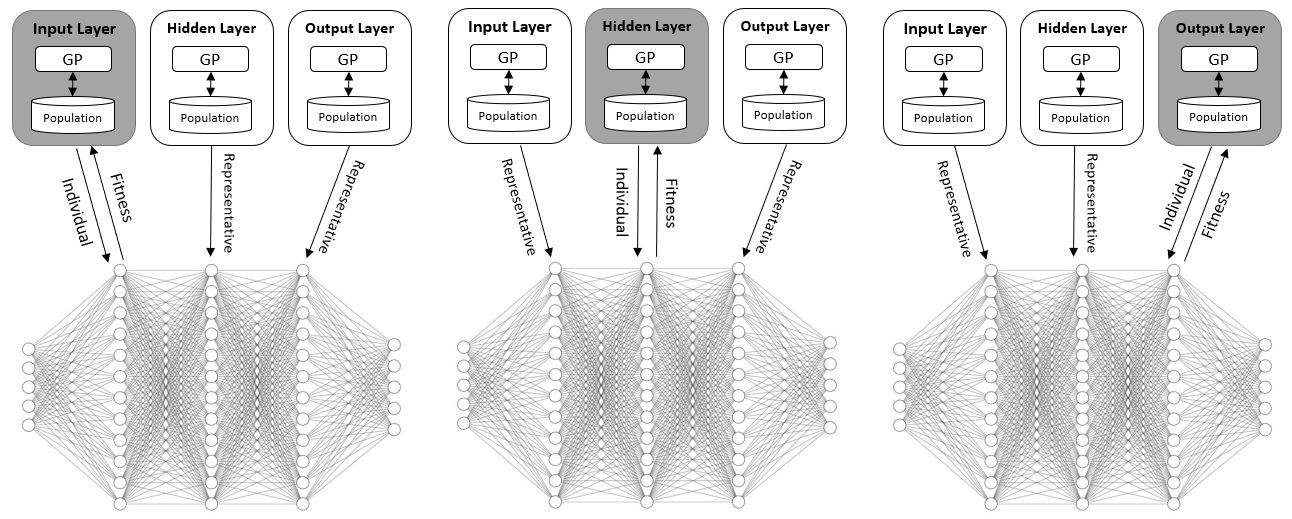}\hfill
    \caption{Fitness computation of an individual in Coevo requires cooperation from the other two populations.}
    \vspace{0.1cm}
    \label{fig:coevo}
\end{figure*}

Our coevolutionary algorithm comprises three separate populations: 1) input-layer AFs, 2) hidden-layer AFs, 3) output-layer AFs. Combining three individuals---one from each population---results in an AF architecture that can be evaluated.  For each evolved individual in each population---an AF---we evaluated its fitness by choosing the two best (top-fitness) individuals of the two other populations from the previous generation. Each population evolved for $max\_iter=50$ iterations (generations). We then created a neural network composed of the three AFs in the respective input, hidden, and output layers. 
The scheme for fitness computation is depicted in Figure~\ref{fig:coevo}.
As with Evo-Single and Evo-Triple, training uses the train$_1$ set, and fitness is computed over the train$_2$ set.
The fitness-computation scheme is shown in Figure~\ref{fig:coevo}.
The coevolutionary algorithm is described in Algorithm~\ref{alg:coevolution}. 

\begin{algorithm*}[]
\small
\caption{Coevolutionary Algorithm: Main Components.}
\label{alg:coevolution}
\begin{algorithmic}[1]

\Statex
\Function{create\_network}{\textit{individual}, \textit{best\_input}, \textit{best\_hidden}, \textit{best\_output}, $\mathit{pop}$}  \Comment{$\mathit{pop \in \{input, hidden, output\}}$}
    \If{$\mathit{pop==input}$}
        \State \textit{model} $\leftarrow$ \textsc{NeuralNetwork}\textit{(individual, best\_hidden, best\_output)}
    \ElsIf{$\mathit{pop==hidden}$}
        \State \textit{model} $\leftarrow$ \textsc{NeuralNetwork}\textit{(best\_input, individual, best\_output)}
    \Else \Comment{{$\mathit{pop==output}$}}
        \State \textit{model} $\leftarrow$ \textsc{NeuralNetwork}\textit{(best\_input, best\_hidden, individual)}
    \EndIf
    \Return{model}
\EndFunction

\Statex
\Function{NeuralNetwork}{\textit{input\_af}, \textit{hidden\_af}, \textit{output\_af}}
    \State \textit{model $\leftarrow$} Create a neural network  with the given AFs, and input/output dimensions determined by dataset
    \State \textbf{return} \textit{model}
\EndFunction

\Statex
\Function{fitness}{\textit{model}, \textit{train$_1$}, \textit{train$_2$}}
    \State train model for 3 epochs with cross entropy loss and mini-batches of size $|\mathit{train_1}|$/$10$
    \State \textbf{return} \textit{accuracy score} over \textit{train$_2$}
\EndFunction

\end{algorithmic}
\normalsize
\end{algorithm*}

There might arise a concern over the differentiability of evolved AFs (PyTorch performs automatic differentiation---indeed this feature is one of its main strengths). However, in practice, this does not pose a problem since the number of points at which most functions are non-differentiable is usually (infinitely) small. For example, ReLU is non-differentiable at zero, however, in practice, it is rare to stumble often onto this value, and when we do, an arbitrary value can be assigned (e.g., zero), with little to no effect on performance. Our experiments have not shown differentiability  to be of any concern.

\section{Experimental Setup} 
\label{sec:exp}
We experimented with four supervised classification datasets---MNIST, FashionMNIST, KMNIST, USPS---using PyTorch's \cite{paszke2019pytorch} default data loaders, with pixel values normalized between $[0,1]$ in the preprocessing step. The architectures used in the experiments are fixed and delineated in Tables~\ref{tab:ann_params} and~\ref{tab:cnn_params}. Full information about the datasets is shown in Table~\ref{tab:datasets}. 

We used the Torchvision library---an accessible, open-source, machine-vision package for PyTorch, written in C++ \cite{10.1145/1873951.1874254}. We used our GPU cluster\footnote{Nvidia GPU cards: 19 RTX 3090, 56 RTX 2080, 52 GTX 1080, 2 Titan XP, 4 P100.}, performing 20 replicate runs for each experimental setup. We split the data three ways into train$_1$ (60\%, used to ``locally'' train the networks), train$_2$ (25\%, used to ``globally'' evolve the networks), and test (15\%) sets, using the training sets for the different automated approaches, while saving the test set for post-evaluation. 

\begin{table}
\centering
\caption{Datasets. Note: 1 $\times$ $h$ $\times$ $w$ represents a single-channel (greyscale) image of height $h$ and width $w$.}
\label{tab:datasets}
\begin{tabular}{rcccl}
 Dataset & Images & Classes & Training & Test  \\ \hline
 MNIST & 1 $\times$ 28 $\times$ 28 & 10 & 60,000 & 10,000 \\
 KMNIST & 1 $\times$ 28 $\times$ 28 & 10 & 60,000 & 10,000 \\
 FashionMNIST & 1 $\times$ 28 $\times$ 28 & 10 & 60,000 & 10,000 \\
 USPS & 1 $\times$ 16 $\times$ 16 & 10 & 7291 & 2007 \\
\end{tabular}
\end{table}

As noted above, we compared 5 different approaches: Standard-FCN/CNN, Random-FCN/CNN, Evo-Single, Evo-Triple, and Coevo. 
We ran multiple replicates of four kinds, each involving the execution and comparison of five different methods:
\begin{enumerate}
    \item FCN, ReLU: Run Standard-FCN with ReLU, Random-FCN, Evo-Single, Evo-Triple, Coevo.
    \item FCN, Leaky ReLU: Run Standard-FCN with Leaky ReLU, Random-FCN, Evo-Single, Evo-Triple, Coevo.
    \item CNN, ReLU: Run Standard-CNN with ReLU, Random-CNN, Evo-Single, Evo-Triple, Coevo.
    \item CNN, Leaky ReLU: Run Standard-CNN with Leaky ReLU, Random-CNN, Evo-Single, Evo-Triple, Coevo.
\end{enumerate}

The score of a network at any given phase was its classification accuracy over said phase's dataset.
Scoring a standard network (either FCN or CNN) was done as follows: train the network using as training set both the train$_1$ and train$_2$ sets; the trained network's final score for comparison purposes was its performance over the test set.
Scoring the other types of networks (random, evolved, and coevolved) was done as follows:
train the network using the train$_1$ set; use as fitness the score over the train$_2$ set;
the network's final score for comparison purposes was its performance over the test set.
Note that the random and evolutionary algorithms never saw the test set---only the train$_1$ and train$_2$ sets were provided; solely the best individual returned was ascribed a test score based on the test set. In the next section, wherein we describe our results, values shown are   test scores (i.e., over test sets).

Algorithm~\ref{alg:setup} delineates the experimental setup in pseudo-code format.

\begin{algorithm}
\small
\caption{Experimental Setup}
\label{alg:setup}
\begin{algorithmic}[1]
\Statex
\Require
\Indent
\Statex \textit{dataset} $\gets$ dataset to be used
\Statex \textit{replicates} $\gets$ number of replicates to run
\EndIndent

\Ensure
\Indent
\Statex Performance measures (over test sets)
\EndIndent

\Statex

\For{\textit{rep} \textbf{in} \textit{\{1\ldots replicates\}}}
    \State Shuffle and split the dataset into \textit{train$_1$ set}, \textit{train$_2$ set}, \textit{test set}
    \State \textit{Standard-Net} $\gets$ create a standard neural network \Comment{\textit{Net} is either FCN or CNN}
    \State \textit{Random-Net} $\gets$ best network obtained through random search
    \State \textit{Evo-Single-Net} $\gets$ best network obtained through Evo-Single method
    \State \textit{Evo-Triple-Net} $\gets$ best network obtained through Evo-Triple method
    \State \textit{Coevo-Net} $\gets$ best network obtained through coevolution (Algorithm~\ref{alg:coevolution})
    \For {\textit{net} \textbf{in} \textit{\{Standard-Net, Random-Net, Evo-Single-Net, Evo-Triple-Net, Coevo-Net\}}}
          \State Train \textit{net} on \textit{train$_1$ + train$_2$ sets} for 30 epochs
          \State Test \textit{net} on \textit{test set} 
          \State Record \textit{net}'s \textit{test} performance 
    \EndFor
\EndFor

\end{algorithmic}
\normalsize
\end{algorithm}

\begin{figure*}
\centering
\includegraphics[width=0.7\textwidth]{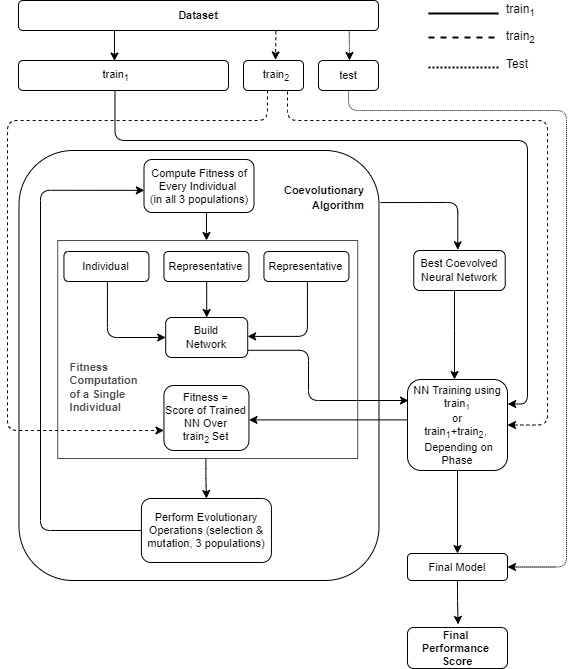}
\caption{Overview of experimental coevolutionary setup.
The coevolutionary algorithm uses the train$_1$ set to train a model and the train$_2$ set to evaluate the evolved model's fitness. The train$_1$ and train$_2$ sets are combined together and used as the training set for the best coevolved neural network. The final model is tested over the test set.}
\label{fig:flow_diagram}
\end{figure*}

Figure~\ref{fig:flow_diagram} depicts an overview of the coevolutionary experimental setup.
The code is available at \url{https://github.com/razla}.

\section{Results and Analysis}
\label{sec:res}

Our experimental results are shown in Table~\ref{tab:results}. The mean of the highest per-replicate test accuracy score over all replicates, for each method, is reported as the best score for that method. 

\begin{table*}
\caption{Experimental results. 
Each line represents 20 replicates.
Values shown are mean accuracy percentages over test set, where the mean is computed over best per-replicate scores for a particular method. The winning method is marked in boldface.
Leaky: Leaky ReLU; Arch: Architecture.
}    
\label{tab:results}
    \begin{center}
        \resizebox{0.7\textwidth}{!}{
            \begin{tabular}{ |c|c|c|c|c|c|c|c| }
                \hline
                \textbf{Dataset} & \textbf{Baseline} & \textbf{Arch} & \textbf{Standard} & \textbf{Random} & \textbf{Evo-Single} & \textbf{Evo-Triple} & \textbf{Coevo} \\
                \hline
                \textbf{\multirow{4}{*}{MNIST}} & \multirow{2}{*}{ReLU} & FCN & 93.93 & 94.48 & 93.85 & 94.13 & \bb{95.9} \\ \cline{3-8}
                 & \multirow{4}{*}{Leaky} & CNN & 96.88 & 99.13 & 98.93 & \bb{99.3} & 99.0 \\ \cline{2-8}
                 & & FCN & 94.98 & 93.68 & 93.53 & \bb{95.7} & 95.5 \\ \cline{3-8}
                 & & CNN & 99.05 & 98.88 & 98.63 & 99.13 & \bb{99.3} \\ 
                \hline
                \textbf{\multirow{4}{*}{KMNIST}} & \multirow{2}{*}{ReLU} & FCN & 80.3 & 80.58 & 80.88 & \bb{81.33} & 80.15 \\ \cline{3-8}
                 & \multirow{4}{*}{Leaky} & CNN & 59.58 & \bb{95.15} & 94.78 & 94.68 & 94.8 \\ \cline{2-8}
                 & & FCN & 78.33 & 80.15 & 81.23 & 80.45 & \bb{81.68} \\ \cline{3-8}
                 & & CNN & 94.7 & 93.98 & 93.15 & \bb{95.38} & 94.23 \\
                \hline
                \textbf{\multirow{4}{*}{FashionMNIST}} & \multirow{2}{*}{ReLU} & FCN & 85.55 & 85.78 & 83.25 & 86.0 & \bb{86.48} \\ \cline{3-8}
                 & \multirow{4}{*}{Leaky} & CNN & 88.45 & 89.25 & 89.08 & 84.58 & \bb{90.0} \\ \cline{2-8}
                 & & FCN & 86.25 & 85.58 & 83.13 & \bb{86.43} & 85.68 \\ \cline{3-8}
                 & & CNN & \bb{89.75} & 89.35 & 88.8 & 89.68 & 89.13 \\
                \hline
                \textbf{\multirow{4}{*}{USPS}} & \multirow{2}{*}{ReLU} & FCN & 90.52 & 91.27 & 91.39 & 84.91 & \bb{91.4} \\ \cline{3-8}
                 & \multirow{4}{*}{Leaky} & CNN & 89.1 & \bb{96.13} & 93.77 & 95.39 & 96.01 \\ \cline{2-8}
                 & & FCN & 92.52 & 91.77 & 92.27 & \bb{92.89} & 92.39 \\ \cline{3-8}
                 & & CNN & 96.38 & 95.39 & 94.39 & 96.01 & \bb{96.88} \\
                \hline
            \end{tabular}}
    \end{center}
\end{table*}

Of the 16 groups of runs, Coevo won 7, Evo-Triple -- 6, Random -- 2, Standard -- 1, and Evo-Single -- 0 (wins are sometimes by a small margin but this is usual in this field). 
The standard FCN or CNN network was outperformed by the other methods in 15 out of the 16 groups of experiments, implying that automated AF search methods may increase model performance.

The fact that most wins involved 3 different AFs---either in Evo-Triple or in Coevo---affirms our surmise in Section~\ref{sec:intro}: Combining different AFs improves network performance.

Table~\ref{tab:best_models} presents the top-3 coevolved AF configurations for each of the four datasets. We note that some of the coevolved configurations are simpler and thus easier to use both for the feed-forward phase and the backpropagation phase. For brevity, we presented only the top-3 configurations, although there were many configurations that were composed only of simple evolved functions such as $x$, $2x$, and $x^2$.

\begin{table*}
\centering
\caption{Coevolution: Top-3 AF configurations per dataset, per architecture.
         Each AF is of the form $f(x)=...$ with the $f(x)$ omitted for brevity. Each AF configuration is of the form \{input-layer AF, hidden-layer AF, output-layer AF\}.}
\label{tab:best_models}

\resizebox{0.6\textwidth}{!}{%
\begin{tabular}{rl}
       \multicolumn{2}{c}{MNIST} \\
       \hline
       \multirow{3}{*}{FCN} & \{$min(x, x^2) - max(x^2, \texttt{ReLU}(x))$, $\texttt{RReLU}(x)$, $\texttt{ELU}(x)$\} \\
        & \{$x-max(x^2, \texttt{ReLU}(x))$, $x$, $x$\} \\
                                [1pt]
                                & \{$x^2$, $\texttt{RReLU}(\texttt{RReLU}(\texttt{RReLU}(x)))$, $2x$\} \\
                               
       \hline
       \multirow{3}{*}{CNN} & \{$x-\texttt{CELU}(x)$, $\texttt{Hardswish}(\texttt{ReLU}(x))$, $\texttt{RReLU}(\texttt{ReLU}(\texttt{Hardshrink}(x)))$\} \\
                                & \{$\texttt{Softshrink}(\texttt{Hardswish}(\texttt{Hardshrink}(x)))$, $\texttt{RReLU}(x)$, $\texttt{ReLU}(x)$\} \\[1pt]
                                & \{$\texttt{Hardtanh}(x) - \texttt{CELU}(x)$, $\texttt{Hardswish}(\texttt{RReLU}(\texttt{ReLU}(x)))$, $\texttt{tanh}(x^2)$\} \\
       \hline
\end{tabular}
}

\vspace{5pt}

\resizebox{0.65\textwidth}{!}{%
\begin{tabular}{rl}
       \multicolumn{2}{c}{KMNIST} \\
       \hline
       \multirow{3}{*}{FCN} & \{$\texttt{ELU}(\texttt{Hardswish}(x))$, $\texttt{CELU}(x)$, $\texttt{ReLU}(x)$\} \\
                                & \{$max(0, \texttt{ELU}(x))$, $x$, $\texttt{ELU}(x)$\} \\
                                & \{$\texttt{RReLU}(\texttt{ReLU}(\texttt{Hardshrink}(x)))$, $min(\texttt{Softshrink}(\texttt{ELU}(x)), x)$, $\texttt{ELU}(x)$ \\
       \hline
       \multirow{3}{*}{CNN} & \{$x$, $\texttt{Softshrink}(\texttt{Hardtanh}(x))$, $\texttt{RReLU}(x) - \texttt{Hardswish}(x)$\} \\
                                & \{$\texttt{Hardshrink}(x)$, $\texttt{Softshrink}(\texttt{Hardshrink}(x))$, $\texttt{LeakyRelu}(\texttt{tanh}(x^2))$\} \\
                                & \{$min((\texttt{Hardshrink}(x) * x), \texttt{ELU}(x))$, $\texttt{Softshrink}(\texttt{Hardtanh}(x))
$, $\texttt{Hardtanh}(\texttt{Hardswish}(x))$\}  \\
       \hline
\end{tabular}
}

\vspace{5pt}

\resizebox{0.7\textwidth}{!}{%
\begin{tabular}{rl}
       \multicolumn{2}{c}{FashionMNIST} \\
       \hline
       \multirow{3}{*}{FCN} & \{$\texttt{RReLU}(x) \cdot  x + x$, $\texttt{ELU}(x)$, $2x$\} \\
                                & \{$x^2 + x$, $max(\texttt{ReLU}(x), \texttt{LeakyRelu}(x) + 2x) - (\texttt{LeakyRelu}(x) + 2x)$, $2x$\} \\
                                & \{$\texttt{ELU}(max(\texttt{ReLU}(x), x))$, $\texttt{CELU}(\texttt{RReLU}(x))$, $\texttt{RReLU}(x)$\}  \\
       \hline
       \multirow{3}{*}{CNN} & \{$\texttt{Hardswish}(x)$, $\texttt{RReLU}(x)$, $\texttt{ELU}(\texttt{LeakyRelu}(x) - x))$\} \\
                                & \{$\texttt{Hardswish}(x)$, $\texttt{Hardswish}(x)$, $\texttt{Softshrink}(\texttt{Softshrink}(\texttt{CELU}(x))$\} \\
                                & \{$\texttt{Softshrink}(x)$, $\texttt{Hardtanh}(\texttt{LeakyRelu}(\texttt{Hardswish}(x)))$, $\texttt{Hardtanh}(max(\texttt{Hardtanh}(\texttt{Hardswish}(x)), 0)$\}  \\
       \hline
\end{tabular}
}

\vspace{5pt}

\resizebox{0.7\textwidth}{!}{%
\begin{tabular}{rl}
       \multicolumn{2}{c}{USPS} \\
       \hline
       \multirow{3}{*}{FCN} & \{$\texttt{ELU}(\texttt{Hardswish}(x))$, $\texttt{CELU}(x)$, $\texttt{ReLU}(x)$\} \\
                                & \{$\texttt{Hardswish}(x)$, $x$, $\texttt{ELU}(x)$\} \\
                                & \{$\texttt{ReLU}(\texttt{CELU}(x))$, $\texttt{Hardswish}(\texttt{tanh}(x)) + \texttt{RReLU}(x)$, $\texttt{LeakyReLU}(\texttt{ELU}(x))$\} \\
       \hline
       \multirow{3}{*}{CNN} &  \{$\texttt{RReLU}(\texttt{tanh}(x))$, $\texttt{tanh}(x)$, $\texttt{LeakyReLU}(x)$\}\\
                                 &  \{$\texttt{HardSwish}(\texttt{Softshrink}(x))$, $\texttt{CELU}(\texttt{ELU}(x) - min(\texttt{LeakyReLU}(x), x))$, $min(\texttt{Hardshrink}(x) - x, \texttt{LeakyReLU}(x))$\}   \\
                                 &  \{$\texttt{HardSwish}(\texttt{Softshrink}(x))$, $\texttt{ELU}(x) - min(\texttt{LeakyReLU}(x), x)$, $\texttt{Softshrink}(min(\texttt{Softshrink}(x), \texttt{Hardshrink}(x) - x))
$\}   \\
\hline
\end{tabular}
}

\end{table*}

\section{Concluding Remarks}
\label{sec:remarks}
Despite the fact that deep-learning techniques are used for a wide range of applications, many hyperparameters must still be manually specified. The choice of AFs is a critical element that is often overlooked (i.e., standard AFs are used). We investigated the significance of AFs in learning in FCN and CNN models for image classification tasks. We introduced a coevolutionary algorithm to evolve new AFs and combine them beneficially, showing that our method performs well on four benchmark datasets. 

We suggest a number of avenues for future research:
\begin{itemize}
    \item Our focus herein was on classification tasks. Other deep-learning tasks can be examined too.
    \item During DNN learning, the influence of AFs on computational complexity could be analyzed.
    \item Applying coevolutionary technique to other DNN hyperparameter searches.
    \item Inspecting other DNN topologies, such as recurrent neural networks (RNNs) and autoencoders (AEs).
\end{itemize}

\clearpage
\bibliographystyle{ACM-Reference-Format}
\bibliography{refs}


\begin{thebibliography}{32}


\ifx \showCODEN    \undefined \def \showCODEN     #1{\unskip}     \fi
\ifx \showDOI      \undefined \def \showDOI       #1{#1}\fi
\ifx \showISBNx    \undefined \def \showISBNx     #1{\unskip}     \fi
\ifx \showISBNxiii \undefined \def \showISBNxiii  #1{\unskip}     \fi
\ifx \showISSN     \undefined \def \showISSN      #1{\unskip}     \fi
\ifx \showLCCN     \undefined \def \showLCCN      #1{\unskip}     \fi
\ifx \shownote     \undefined \def \shownote      #1{#1}          \fi
\ifx \showarticletitle \undefined \def \showarticletitle #1{#1}   \fi
\ifx \showURL      \undefined \def \showURL       {\relax}        \fi
\providecommand\bibfield[2]{#2}
\providecommand\bibinfo[2]{#2}
\providecommand\natexlab[1]{#1}
\providecommand\showeprint[2][]{arXiv:#2}

\bibitem[\protect\citeauthoryear{Agostinelli, Hoffman, Sadowski, and
  Baldi}{Agostinelli et~al\mbox{.}}{2014}]%
        {agostinelli2014learning}
\bibfield{author}{\bibinfo{person}{Forest Agostinelli},
  \bibinfo{person}{Matthew Hoffman}, \bibinfo{person}{Peter Sadowski}, {and}
  \bibinfo{person}{Pierre Baldi}.} \bibinfo{year}{2014}\natexlab{}.
\newblock \showarticletitle{Learning activation functions to improve deep
  neural networks}.
\newblock \bibinfo{journal}{\emph{arXiv preprint arXiv:1412.6830}}
  (\bibinfo{year}{2014}).
\newblock


\bibitem[\protect\citeauthoryear{Basirat and Roth}{Basirat and Roth}{2018}]%
        {basirat2018quest}
\bibfield{author}{\bibinfo{person}{Mina Basirat} {and} \bibinfo{person}{Peter~M
  Roth}.} \bibinfo{year}{2018}\natexlab{}.
\newblock \showarticletitle{The quest for the golden activation function}.
\newblock \bibinfo{journal}{\emph{arXiv preprint arXiv:1808.00783}}
  (\bibinfo{year}{2018}).
\newblock


\bibitem[\protect\citeauthoryear{Bergstra and Bengio}{Bergstra and
  Bengio}{2012}]%
        {bergstra2012random}
\bibfield{author}{\bibinfo{person}{James Bergstra} {and}
  \bibinfo{person}{Yoshua Bengio}.} \bibinfo{year}{2012}\natexlab{}.
\newblock \showarticletitle{Random search for hyper-parameter optimization.}
\newblock \bibinfo{journal}{\emph{Journal of machine learning research}}
  \bibinfo{volume}{13}, \bibinfo{number}{2} (\bibinfo{year}{2012}).
\newblock


\bibitem[\protect\citeauthoryear{Bingham, Macke, and Miikkulainen}{Bingham
  et~al\mbox{.}}{2020}]%
        {10.1145/3377930.3389841}
\bibfield{author}{\bibinfo{person}{Garrett Bingham}, \bibinfo{person}{William
  Macke}, {and} \bibinfo{person}{Risto Miikkulainen}.}
  \bibinfo{year}{2020}\natexlab{}.
\newblock \showarticletitle{Evolutionary Optimization of Deep Learning
  Activation Functions}. In \bibinfo{booktitle}{\emph{Proceedings of the 2020
  Genetic and Evolutionary Computation Conference}} (Canc\'{u}n, Mexico)
  \emph{(\bibinfo{series}{GECCO '20})}. \bibinfo{publisher}{Association for
  Computing Machinery}, \bibinfo{address}{New York, NY, USA},
  \bibinfo{pages}{289–296}.
\newblock
\showISBNx{9781450371285}


\bibitem[\protect\citeauthoryear{Bingham and Miikkulainen}{Bingham and
  Miikkulainen}{2022}]%
        {bingham2022discovering}
\bibfield{author}{\bibinfo{person}{Garrett Bingham} {and}
  \bibinfo{person}{Risto Miikkulainen}.} \bibinfo{year}{2022}\natexlab{}.
\newblock \showarticletitle{Discovering parametric activation functions}.
\newblock \bibinfo{journal}{\emph{Neural Networks}} (\bibinfo{year}{2022}).
\newblock


\bibitem[\protect\citeauthoryear{Breuel}{Breuel}{2015}]%
        {breuel2015effects}
\bibfield{author}{\bibinfo{person}{Thomas~M Breuel}.}
  \bibinfo{year}{2015}\natexlab{}.
\newblock \showarticletitle{The effects of hyperparameters on {SGD} training of
  neural networks}.
\newblock \bibinfo{journal}{\emph{arXiv preprint arXiv:1508.02788}}
  (\bibinfo{year}{2015}).
\newblock


\bibitem[\protect\citeauthoryear{Ding, Wu, Huang, Tang, Wu, Yang, Zhu, and
  Zhuang}{Ding et~al\mbox{.}}{2022}]%
        {DING2022}
\bibfield{author}{\bibinfo{person}{Yadong Ding}, \bibinfo{person}{Yu Wu},
  \bibinfo{person}{Chengyue Huang}, \bibinfo{person}{Siliang Tang},
  \bibinfo{person}{Fei Wu}, \bibinfo{person}{Yi Yang}, \bibinfo{person}{Wenwu
  Zhu}, {and} \bibinfo{person}{Yueting Zhuang}.}
  \bibinfo{year}{2022}\natexlab{}.
\newblock \showarticletitle{NAP: Neural Architecture search with Pruning}.
\newblock \bibinfo{journal}{\emph{Neurocomputing}} (\bibinfo{year}{2022}).
\newblock
\showISSN{0925-2312}
\urldef\tempurl%
\url{https://doi.org/10.1016/j.neucom.2021.12.002}
\showDOI{\tempurl}


\bibitem[\protect\citeauthoryear{Franchini, Ruggiero, Porta, and
  Zanni}{Franchini et~al\mbox{.}}{2023}]%
        {Franchini2023}
\bibfield{author}{\bibinfo{person}{Giorgia Franchini}, \bibinfo{person}{Valeria
  Ruggiero}, \bibinfo{person}{Federica Porta}, {and} \bibinfo{person}{Luca
  Zanni}.} \bibinfo{year}{2023}\natexlab{}.
\newblock \showarticletitle{Neural architecture search via standard machine
  learning methodologies}.
\newblock \bibinfo{journal}{\emph{Mathematics in Engineering}}
  (\bibinfo{year}{2023}).
\newblock
\urldef\tempurl%
\url{https://www.aimspress.com/article/doi/10.3934/mine.2023012?viewType=HTML}
\showURL{%
\tempurl}


\bibitem[\protect\citeauthoryear{Hochreiter}{Hochreiter}{1998}]%
        {hochreiter1998vanishing}
\bibfield{author}{\bibinfo{person}{Sepp Hochreiter}.}
  \bibinfo{year}{1998}\natexlab{}.
\newblock \showarticletitle{The vanishing gradient problem during learning
  recurrent neural nets and problem solutions}.
\newblock \bibinfo{journal}{\emph{International Journal of Uncertainty,
  Fuzziness and Knowledge-Based Systems}} \bibinfo{volume}{6},
  \bibinfo{number}{02} (\bibinfo{year}{1998}), \bibinfo{pages}{107--116}.
\newblock


\bibitem[\protect\citeauthoryear{Itano, de~Abreu~de Sousa, and
  Del-Moral-Hernandez}{Itano et~al\mbox{.}}{2018}]%
        {8489520}
\bibfield{author}{\bibinfo{person}{Fernando Itano},
  \bibinfo{person}{Miguel~Angelo de~Abreu~de Sousa}, {and}
  \bibinfo{person}{Emilio Del-Moral-Hernandez}.}
  \bibinfo{year}{2018}\natexlab{}.
\newblock \showarticletitle{Extending {MLP} {ANN} hyper-parameters Optimization
  by using Genetic Algorithm}. In \bibinfo{booktitle}{\emph{2018 International
  Joint Conference on Neural Networks (IJCNN)}}. \bibinfo{pages}{1--8}.
\newblock


\bibitem[\protect\citeauthoryear{Lu, Shin, Su, and Karniadakis}{Lu
  et~al\mbox{.}}{2019}]%
        {lu2019dying}
\bibfield{author}{\bibinfo{person}{Lu Lu}, \bibinfo{person}{Yeonjong Shin},
  \bibinfo{person}{Yanhui Su}, {and} \bibinfo{person}{George~Em Karniadakis}.}
  \bibinfo{year}{2019}\natexlab{}.
\newblock \showarticletitle{Dying {ReLU} and initialization: Theory and
  numerical examples}.
\newblock \bibinfo{journal}{\emph{arXiv preprint arXiv:1903.06733}}
  (\bibinfo{year}{2019}).
\newblock


\bibitem[\protect\citeauthoryear{Marcel and Rodriguez}{Marcel and
  Rodriguez}{2010}]%
        {10.1145/1873951.1874254}
\bibfield{author}{\bibinfo{person}{S\'{e}bastien Marcel} {and}
  \bibinfo{person}{Yann Rodriguez}.} \bibinfo{year}{2010}\natexlab{}.
\newblock \showarticletitle{Torchvision the Machine-Vision Package of {Torch}}.
  In \bibinfo{booktitle}{\emph{Proceedings of the 18th ACM International
  Conference on Multimedia}}. \bibinfo{publisher}{Association for Computing
  Machinery}, \bibinfo{address}{New York, NY, USA},
  \bibinfo{pages}{1485–1488}.
\newblock
\showISBNx{9781605589336}


\bibitem[\protect\citeauthoryear{Miikkulainen, Liang, Meyerson, Rawal, Fink,
  Francon, Raju, Shahrzad, Navruzyan, Duffy, et~al\mbox{.}}{Miikkulainen
  et~al\mbox{.}}{2019}]%
        {miikkulainen2019evolving}
\bibfield{author}{\bibinfo{person}{Risto Miikkulainen}, \bibinfo{person}{Jason
  Liang}, \bibinfo{person}{Elliot Meyerson}, \bibinfo{person}{Aditya Rawal},
  \bibinfo{person}{Daniel Fink}, \bibinfo{person}{Olivier Francon},
  \bibinfo{person}{Bala Raju}, \bibinfo{person}{Hormoz Shahrzad},
  \bibinfo{person}{Arshak Navruzyan}, \bibinfo{person}{Nigel Duffy},
  {et~al\mbox{.}}} \bibinfo{year}{2019}\natexlab{}.
\newblock \showarticletitle{Evolving deep neural networks}.
\newblock In \bibinfo{booktitle}{\emph{Artificial intelligence in the age of
  neural networks and brain computing}}. \bibinfo{publisher}{Elsevier},
  \bibinfo{pages}{293--312}.
\newblock


\bibitem[\protect\citeauthoryear{Miller and Harding}{Miller and
  Harding}{2008}]%
        {10.1145/1388969.1389075}
\bibfield{author}{\bibinfo{person}{Julian~Francis Miller} {and}
  \bibinfo{person}{Simon~L. Harding}.} \bibinfo{year}{2008}\natexlab{}.
\newblock \showarticletitle{Cartesian Genetic Programming}. In
  \bibinfo{booktitle}{\emph{Proceedings of the 10th Annual Conference Companion
  on Genetic and Evolutionary Computation}} (Atlanta, GA, USA)
  \emph{(\bibinfo{series}{GECCO '08})}. \bibinfo{publisher}{Association for
  Computing Machinery}, \bibinfo{address}{New York, NY, USA},
  \bibinfo{pages}{2701–2726}.
\newblock
\showISBNx{9781605581316}


\bibitem[\protect\citeauthoryear{Miller and Smith}{Miller and Smith}{2006}]%
        {miller2006redundancy}
\bibfield{author}{\bibinfo{person}{Julian~F Miller} {and}
  \bibinfo{person}{Stephen~L Smith}.} \bibinfo{year}{2006}\natexlab{}.
\newblock \showarticletitle{Redundancy and computational efficiency in
  {Cartesian} genetic programming}.
\newblock \bibinfo{journal}{\emph{IEEE Transactions on Evolutionary
  Computation}} \bibinfo{volume}{10}, \bibinfo{number}{2}
  (\bibinfo{year}{2006}), \bibinfo{pages}{167--174}.
\newblock


\bibitem[\protect\citeauthoryear{Miller, Thomson, and Fogarty}{Miller
  et~al\mbox{.}}{1997}]%
        {miller1997designing}
\bibfield{author}{\bibinfo{person}{Julian~F Miller}, \bibinfo{person}{Peter
  Thomson}, {and} \bibinfo{person}{Terence Fogarty}.}
  \bibinfo{year}{1997}\natexlab{}.
\newblock \showarticletitle{Designing electronic circuits using evolutionary
  algorithms. arithmetic circuits: A case study}.
\newblock \bibinfo{journal}{\emph{Genetic Algorithms and Evolution Strategies
  in Engineering and Computer Science}} (\bibinfo{year}{1997}),
  \bibinfo{pages}{105--131}.
\newblock


\bibitem[\protect\citeauthoryear{Nader and Azar}{Nader and Azar}{2021}]%
        {Nader2021}
\bibfield{author}{\bibinfo{person}{Andrew Nader} {and}
  \bibinfo{person}{Danielle Azar}.} \bibinfo{year}{2021}\natexlab{}.
\newblock \showarticletitle{Evolution of Activation Functions: An Empirical
  Investigation}.
\newblock \bibinfo{journal}{\emph{ACM Transactions on Evolutionary Learning and
  Optimization}} \bibinfo{volume}{1}, \bibinfo{number}{2}
  (\bibinfo{year}{2021}), \bibinfo{pages}{1--36}.
\newblock


\bibitem[\protect\citeauthoryear{Nwankpa, Ijomah, Gachagan, and
  Marshall}{Nwankpa et~al\mbox{.}}{2018}]%
        {nwankpa2018activation}
\bibfield{author}{\bibinfo{person}{Chigozie Nwankpa}, \bibinfo{person}{Winifred
  Ijomah}, \bibinfo{person}{Anthony Gachagan}, {and} \bibinfo{person}{Stephen
  Marshall}.} \bibinfo{year}{2018}\natexlab{}.
\newblock \showarticletitle{Activation functions: Comparison of trends in
  practice and research for deep learning}.
\newblock \bibinfo{journal}{\emph{arXiv preprint arXiv:1811.03378}}
  (\bibinfo{year}{2018}).
\newblock


\bibitem[\protect\citeauthoryear{Paszke, Gross, Massa, Lerer, Bradbury, Chanan,
  Killeen, Lin, Gimelshein, Antiga, et~al\mbox{.}}{Paszke
  et~al\mbox{.}}{2019}]%
        {paszke2019pytorch}
\bibfield{author}{\bibinfo{person}{Adam Paszke}, \bibinfo{person}{Sam Gross},
  \bibinfo{person}{Francisco Massa}, \bibinfo{person}{Adam Lerer},
  \bibinfo{person}{James Bradbury}, \bibinfo{person}{Gregory Chanan},
  \bibinfo{person}{Trevor Killeen}, \bibinfo{person}{Zeming Lin},
  \bibinfo{person}{Natalia Gimelshein}, \bibinfo{person}{Luca Antiga},
  {et~al\mbox{.}}} \bibinfo{year}{2019}\natexlab{}.
\newblock \showarticletitle{{PyTorch}: An imperative style, high-performance
  deep learning library}.
\newblock \bibinfo{journal}{\emph{Advances in neural information processing
  systems}}  \bibinfo{volume}{32} (\bibinfo{year}{2019}),
  \bibinfo{pages}{8026--8037}.
\newblock


\bibitem[\protect\citeauthoryear{Pena-Reyes and Sipper}{Pena-Reyes and
  Sipper}{2001}]%
        {Pena:2001}
\bibfield{author}{\bibinfo{person}{Carlos~Andr{\'e}s Pena-Reyes} {and}
  \bibinfo{person}{Moshe Sipper}.} \bibinfo{year}{2001}\natexlab{}.
\newblock \showarticletitle{Fuzzy {CoCo}: A cooperative-coevolutionary approach
  to fuzzy modeling}.
\newblock \bibinfo{journal}{\emph{IEEE Transactions on Fuzzy Systems}}
  \bibinfo{volume}{9}, \bibinfo{number}{5} (\bibinfo{year}{2001}),
  \bibinfo{pages}{727--737}.
\newblock


\bibitem[\protect\citeauthoryear{Quade}{Quade}{2020}]%
        {Ohjeah}
\bibfield{author}{\bibinfo{person}{Markus Quade}.}
  \bibinfo{year}{2020}\natexlab{}.
\newblock \bibinfo{title}{cartesian}.
\newblock \bibinfo{howpublished}{\url{https://github.com/Ohjeah/cartesian}}.
\newblock


\bibitem[\protect\citeauthoryear{Saha, Nagaraj, Mathur, and Yedida}{Saha
  et~al\mbox{.}}{2019}]%
        {saha2019evolution}
\bibfield{author}{\bibinfo{person}{Snehanshu Saha}, \bibinfo{person}{Nithin
  Nagaraj}, \bibinfo{person}{Archana Mathur}, {and} \bibinfo{person}{Rahul
  Yedida}.} \bibinfo{year}{2019}\natexlab{}.
\newblock \showarticletitle{Evolution of novel activation functions in neural
  network training with applications to classification of exoplanets}.
\newblock \bibinfo{journal}{\emph{arXiv preprint arXiv:1906.01975}}
  (\bibinfo{year}{2019}).
\newblock


\bibitem[\protect\citeauthoryear{Sharma and Sharma}{Sharma and Sharma}{2017}]%
        {sharma2017activation}
\bibfield{author}{\bibinfo{person}{Sagar Sharma} {and} \bibinfo{person}{Simone
  Sharma}.} \bibinfo{year}{2017}\natexlab{}.
\newblock \showarticletitle{Activation functions in neural networks}.
\newblock \bibinfo{journal}{\emph{Towards Data Science}} \bibinfo{volume}{6},
  \bibinfo{number}{12} (\bibinfo{year}{2017}), \bibinfo{pages}{310--316}.
\newblock


\bibitem[\protect\citeauthoryear{Sipper}{Sipper}{2021}]%
        {sipper2021neural}
\bibfield{author}{\bibinfo{person}{Moshe Sipper}.}
  \bibinfo{year}{2021}\natexlab{}.
\newblock \showarticletitle{Neural Networks with {\`A} La Carte Selection of
  Activation Functions}.
\newblock \bibinfo{journal}{\emph{SN Computer Science}} \bibinfo{volume}{2},
  \bibinfo{number}{6} (\bibinfo{year}{2021}), \bibinfo{pages}{1--9}.
\newblock


\bibitem[\protect\citeauthoryear{Sipper, Moore, and Urbanowicz}{Sipper
  et~al\mbox{.}}{2019}]%
        {sipper2019solution}
\bibfield{author}{\bibinfo{person}{Moshe Sipper}, \bibinfo{person}{Jason~H
  Moore}, {and} \bibinfo{person}{Ryan~J Urbanowicz}.}
  \bibinfo{year}{2019}\natexlab{}.
\newblock \showarticletitle{Solution and fitness evolution ({SAFE}): Coevolving
  solutions and their objective functions}. In
  \bibinfo{booktitle}{\emph{European Conference on Genetic Programming}}.
  Springer, \bibinfo{pages}{146--161}.
\newblock


\bibitem[\protect\citeauthoryear{Snoek, Larochelle, and Adams}{Snoek
  et~al\mbox{.}}{2012}]%
        {snoek2012practical}
\bibfield{author}{\bibinfo{person}{Jasper Snoek}, \bibinfo{person}{Hugo
  Larochelle}, {and} \bibinfo{person}{Ryan~P Adams}.}
  \bibinfo{year}{2012}\natexlab{}.
\newblock \showarticletitle{Practical {Bayesian} optimization of machine
  learning algorithms}.
\newblock \bibinfo{journal}{\emph{arXiv preprint arXiv:1206.2944}}
  (\bibinfo{year}{2012}).
\newblock


\bibitem[\protect\citeauthoryear{Stanley and Miikkulainen}{Stanley and
  Miikkulainen}{2002}]%
        {stanley2002evolving}
\bibfield{author}{\bibinfo{person}{Kenneth~O Stanley} {and}
  \bibinfo{person}{Risto Miikkulainen}.} \bibinfo{year}{2002}\natexlab{}.
\newblock \showarticletitle{Evolving neural networks through augmenting
  topologies}.
\newblock \bibinfo{journal}{\emph{Evolutionary computation}}
  \bibinfo{volume}{10}, \bibinfo{number}{2} (\bibinfo{year}{2002}),
  \bibinfo{pages}{99--127}.
\newblock


\bibitem[\protect\citeauthoryear{Sun, Xue, Zhang, and Yen}{Sun
  et~al\mbox{.}}{2019}]%
        {sun2019evolving}
\bibfield{author}{\bibinfo{person}{Yanan Sun}, \bibinfo{person}{Bing Xue},
  \bibinfo{person}{Mengjie Zhang}, {and} \bibinfo{person}{Gary~G Yen}.}
  \bibinfo{year}{2019}\natexlab{}.
\newblock \showarticletitle{Evolving deep convolutional neural networks for
  image classification}.
\newblock \bibinfo{journal}{\emph{IEEE Transactions on Evolutionary
  Computation}} \bibinfo{volume}{24}, \bibinfo{number}{2}
  (\bibinfo{year}{2019}), \bibinfo{pages}{394--407}.
\newblock


\bibitem[\protect\citeauthoryear{Wu, Lin, and Tsai}{Wu et~al\mbox{.}}{2022}]%
        {10.1145/3491396.3506510}
\bibfield{author}{\bibinfo{person}{Meng-Ting Wu}, \bibinfo{person}{Hung-I Lin},
  {and} \bibinfo{person}{Chun-Wei Tsai}.} \bibinfo{year}{2022}\natexlab{}.
\newblock \showarticletitle{A Training-Free Genetic Neural Architecture
  Search}. In \bibinfo{booktitle}{\emph{Proceedings of the 2021 ACM
  International Conference on Intelligent Computing and Its Emerging
  Applications}} (2022-01-01) \emph{(\bibinfo{series}{ACM ICEA '21})}.
  \bibinfo{publisher}{Association for Computing Machinery},
  \bibinfo{address}{Jinan, China}, \bibinfo{pages}{65–70}.
\newblock
\showISBNx{9781450391603}
\urldef\tempurl%
\url{https://doi.org/10.1145/3491396.3506510}
\showDOI{\tempurl}


\bibitem[\protect\citeauthoryear{Zhang, Jin, Jin, and Hao}{Zhang
  et~al\mbox{.}}{2022}]%
        {9672175}
\bibfield{author}{\bibinfo{person}{Haoyu Zhang}, \bibinfo{person}{Yaochu Jin},
  \bibinfo{person}{Yaochu Jin}, {and} \bibinfo{person}{Kuangrong Hao}.}
  \bibinfo{year}{2022}\natexlab{}.
\newblock \showarticletitle{Evolutionary Search for Complete Neural Network
  Architectures with Partial Weight Sharing}.
\newblock \bibinfo{journal}{\emph{IEEE Transactions on Evolutionary
  Computation}} (\bibinfo{year}{2022}), \bibinfo{pages}{1--1}.
\newblock
\urldef\tempurl%
\url{https://doi.org/10.1109/TEVC.2022.3140855}
\showDOI{\tempurl}


\bibitem[\protect\citeauthoryear{Zhang, Zou, and Shi}{Zhang
  et~al\mbox{.}}{2017}]%
        {zhang2017dilated}
\bibfield{author}{\bibinfo{person}{Xiaohu Zhang}, \bibinfo{person}{Yuexian
  Zou}, {and} \bibinfo{person}{Wei Shi}.} \bibinfo{year}{2017}\natexlab{}.
\newblock \showarticletitle{Dilated convolution neural network with {LeakyReLU}
  for environmental sound classification}. In \bibinfo{booktitle}{\emph{2017
  22nd International Conference on Digital Signal Processing (DSP)}}. IEEE,
  \bibinfo{pages}{1--5}.
\newblock


\bibitem[\protect\citeauthoryear{Zhang, Pan, Sun, and Tang}{Zhang
  et~al\mbox{.}}{2018}]%
        {zhang2018multiple}
\bibfield{author}{\bibinfo{person}{Yu-Dong Zhang}, \bibinfo{person}{Chichun
  Pan}, \bibinfo{person}{Junding Sun}, {and} \bibinfo{person}{Chaosheng Tang}.}
  \bibinfo{year}{2018}\natexlab{}.
\newblock \showarticletitle{Multiple sclerosis identification by convolutional
  neural network with dropout and parametric {ReLU}}.
\newblock \bibinfo{journal}{\emph{Journal of computational science}}
  \bibinfo{volume}{28} (\bibinfo{year}{2018}), \bibinfo{pages}{1--10}.
\newblock


\end{thebibliography}

\end{document}